\documentclass[11pt,a4paper]{article}
\usepackage{emnlp2020}
\usepackage{times}
\usepackage{latexsym}
\usepackage{amsmath}
\usepackage{amssymb}
\usepackage{multirow,multicol}
\usepackage{graphicx}

\usepackage{array}
\usepackage{soul}
\usepackage{booktabs}
\usepackage[colorinlistoftodos]{todonotes}

\usepackage{microtype}

\newcommand{\PreserveBackslash}[1]{\let\temp=\\#1\let\\=\temp}
\newcolumntype{C}[1]{>{\PreserveBackslash\centering}p{#1}}
\newcolumntype{R}[1]{>{\PreserveBackslash\raggedleft}p{#1}}
\newcolumntype{L}[1]{>{\PreserveBackslash\raggedright}p{#1}}
\DeclareMathOperator*{\argmax}{\text{argmax}}
\newcommand*\rot{\rotatebox{90}}

\title{COSMIC: COmmonSense knowledge for \\
eMotion Identification in Conversations}

\author{Deepanway Ghosal$^\dagger$,
  Navonil Majumder$^\dagger$,
  Alexander Gelbukh$^\diamond$,\\
  \textbf{Rada Mihalcea$^\triangle$,
  Soujanya Poria$^\dagger$}\\\\

  $^\dagger$ Singapore University of Technology and Design, Singapore\\
  $^\diamond$ CIC, Instituto Polit\'ecnico Nacional, Mexico\\
  $^\triangle$ University of Michigan, USA\\
  \texttt{\{deepanway\_ghosal@mymail.,navonil\_majumder@, sporia@\}sutd.edu.sg},\\ \texttt{gelbukh@cic.ipn.mx, mihalcea@umich.edu} \\
  }

\date{}
\aclfinalcopy
\begin{document}
\maketitle
\begin{abstract}
In this paper, we address the task of utterance level emotion recognition in conversations using commonsense knowledge. We propose {\sc Cosmic}, a new framework that incorporates different elements of commonsense such as mental states, events, and causal relations, and build upon them to learn interactions between interlocutors participating in a conversation. Current state-of-the-art methods often encounter difficulties in context propagation, emotion shift detection, and differentiating between related emotion classes. By learning distinct commonsense representations, {\sc Cosmic} addresses these challenges and achieves new state-of-the-art results for emotion recognition on four different benchmark conversational datasets. Our code is available at \url{https://github.com/declare-lab/conv-emotion}.
\end{abstract}
\begin{figure}[ht!]
    \centering
    \includegraphics[width=\linewidth]{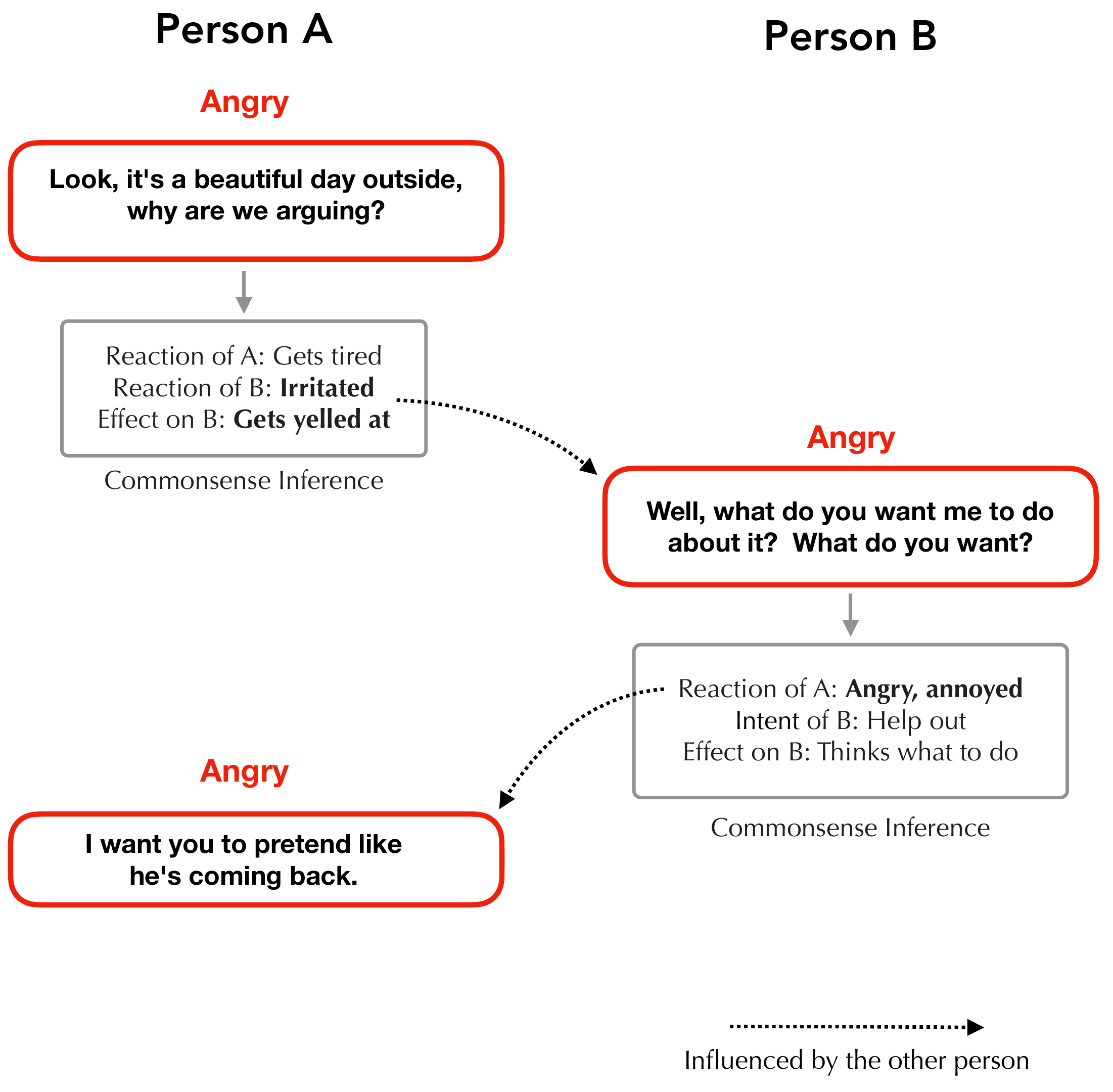}
    \caption{Commonsense knowledge can lead to explainable dialogue understanding. It will help models to understand, reason, and explain events and situations. In this particular example, commonsense inference is applied to a sequence of utterances in a two-party conversation. Person A’s first utterance indicates that he/she is tired of arguing with person B. The tone of the utterance also implies that person B is getting yelled at by person A, which invokes a reaction of irritation in person B. Person B then asks what he/she can do to help and says this while being angry. This again makes person A annoyed and influences him/her to respond with anger. This kind of inferred commonsense knowledge about the reaction, effect, and intent of the speaker and the listener helps in predicting the emotional dynamics of the participants.}
    \label{fig:csk-erc}
\end{figure}

\section{Introduction}
Emotion recognition is a long-standing research problem in Artificial Intelligence (AI). With the growing popularity of conversational AI research, the topic of emotion recognition in conversations has received significant attention from the research community \cite{Li2020MultiTaskLW,ghosal2019dialoguegcn,zhang2019modeling}. Identifying emotions in conversations is a core step toward fine-grained conversation understanding, which in turn is essential for downstream tasks such as emotion-aware chat agents \cite{lin-etal-2019-moel,rashkin2018empathetic}, visual question answering \cite{Tapaswi16,Azab19Multimodal}, health conversations \cite{Althoff16,Perez-Rosas17Predicting} and others.

Natural conversations are complex as they are governed by several distinct variables that affect the flow of a conversation and the emotional dynamics of the participants. These variables include topic, viewpoint, speaker personality, argumentation logic, intent, and so on \cite{poria2019emotion}. 
Additionally, individual utterances are also governed by the mental state, intent, and emotional state of the participants at the time when they are uttered.
In this conversation model,
only the utterances can be observed as the conversation unfolds, while other variables such as speaker state and intent remain latent as they are not directly observed by the other participants. Similarly, the emotional state of the speakers cannot be directly observed, but it can be inferred from the utterances that are observable.\footnote{In multimodal conversations, there are other variables that can be observed, such as facial expressions, gestures, pitch, and acoustic indicators.}

The commonsense knowledge of the participants in a conversation plays a central role in inferring the latent variables of a conversation. It is used to guide the participants through their reasoning about the content of the conversation, dialog planning, decision making, and many other reasoning tasks. It is also used to recognize other finer-grained elements of a conversation, such as avoiding repetition, asking questions, refraining from giving unrelated responses, and so on --- all of which control aspects of the conversation such as fluency, interestingness, inquisitiveness, or empathy. Commonsense knowledge is thus necessary to model the nature and flow of the dialogue and the emotional dynamics of the participants. In Figure \ref{fig:csk-erc}, we illustrate one such scenario where commonsense knowledge is utilized to infer emotions of the utterances in a dialogue.

Natural language is
often indicative of one's emotion. Hence, emotion recognition has
been enjoying popularity in the field of NLP  \citep{kratzwald2018decision, colneric2018emotion}, due to its widespread applications in
opinion mining, recommender systems, healthcare, and so on. Only in the past few years has emotion recognition in conversation (ERC) gained attention from the NLP community~\citep{yeh2019interaction, chen2018emotionlines,dialoguernn,zhou2018emotional}
due to the growing availability of public conversational data. ERC can be used to analyze
conversations that take place on social media. It can also aid in analyzing conversations in real time, which can be instrumental in legal trials, interviews, e-health services, and more. Unlike vanilla emotion recognition of sentences/utterances, ERC ideally requires context
modeling of the individual utterances. This context can be attributed to the preceding
utterances, and relies on the temporal sequence of utterances.
Compared to the recently published works on ERC~\citep{chen2018emotionlines,dialoguernn,zhou2018emotional, qindcr2020dcrnet,zhong2019knowledge,zhang2019modeling}, both lexicon-based~\citep{wu2006emotion,mohammad2010emotions, shaheen2014emotion} and modern deep learning-based~\citep{kratzwald2018decision, colneric2018emotion} vanilla emotion recognition approaches fail to work well on ERC datasets as this work ignores the conversation specific factors such as the presence of contextual cues, the temporality in speakers' turns, or speaker-specific information.

In this paper, we introduce {\sc Cosmic}, a commonsense-guided framework for emotion identification in conversations. By building upon a very large commonsense knowledge base, our proposed framework captures some of the complex interactions between personality, events, mental states, intents, and emotions leading towards a better understanding of the emotional dynamics and other aspects of conversation. Through extensive evaluations on four different conversation datasets and comparisons with several baselines and state-of-the-art models, we show the effectiveness of a model that explicitly accounts for commonsense. Moreover, feature ablation experiments highlight the role that such knowledge plays in identifying emotion in conversations.

\section{Related Work}
\label{sec:related-works}
Emotion recognition has been an active area of research for many years and has been explored across inter-disciplinary fields such as machine learning, signal processing, social and cognitive psychology, etc \cite{picard2010affective}. The seminal work from \citet{ekman1993facial} presented findings on facial expressions, methods to measure facial expression and their relation with human emotion. Acoustic information and visual cues were later used for emotion recognition by \citet{datcu2014semantic}.

However, emotion recognition in conversations has gained popularity only recently due to the emergence of publicly available conversational datasets collected from social media platforms
and scripted situations such as movies and tv-shows \cite{poria2018meld,zahiri2018emotion}.
The main approach towards conversational emotion recognition is to perform contextual modeling in either textual or multimodal setting with deep-learning based algorithms.
\citet{poria-EtAl:2017:Long} used recurrent neural networks for multimodal emotion recognition followed by \cite{dialoguernn}, where party and global states were used for modeling the emotional dynamics. An external knowledge base was used in \cite{zhong2019knowledge} with transformer networks to perform emotion recognition. Some of the other important works include \cite{hazarika2018icon,hazarika-EtAl:2018:N18-1,zadatt,chen2017multimodal,AAAI1817341}. 



\section{Methodology}
\subsection{Task definition}
Given the transcript of a conversation along with speaker information
for each constituent utterance, the ERC task aims to identify the emotion of each utterance from a set of  pre-defined emotions. Figure \ref{fig:csk-erc} illustrates one such
conversation between two people, where each utterance is labeled by the
underlying emotion. Formally, given an input sequence of $N$
utterances $[(u_1, p_1), (u_2,p_2),\dots, (u_N,p_N)]$, where each utterance $u_i=[u_{i,1},u_{i,2},\dots,u_{i,T}]$ consists of $T$ words $u_{i,j}$  spoken by
party $p_i$, the task is to predict the emotion label $e_i$ of
each utterance $u_i$.
\label{sec:method}
In conversational emotion recognition, the task is to classify each of the constituting utterances into its appropriate emotion category. In literature, the main approach towards this problem has been to first produce context independent representations and then perform contextual modeling. We identify these two distinct modeling phases and aim to improve both of them through the proposed {\sc Cosmic} framework.
Our framework consists of three main stages:

\begin{enumerate}
    \item Context independent feature extraction from pretrained transformer language models.
    \item Commonsense feature extraction from a commonsense knowledge graph.
    \item Incorporating  commonsense knowledge to design better contextual representations and using it for the final emotion classification.
\end{enumerate}

\begin{figure*}[ht!]
    \centering
    \includegraphics[width=\linewidth]{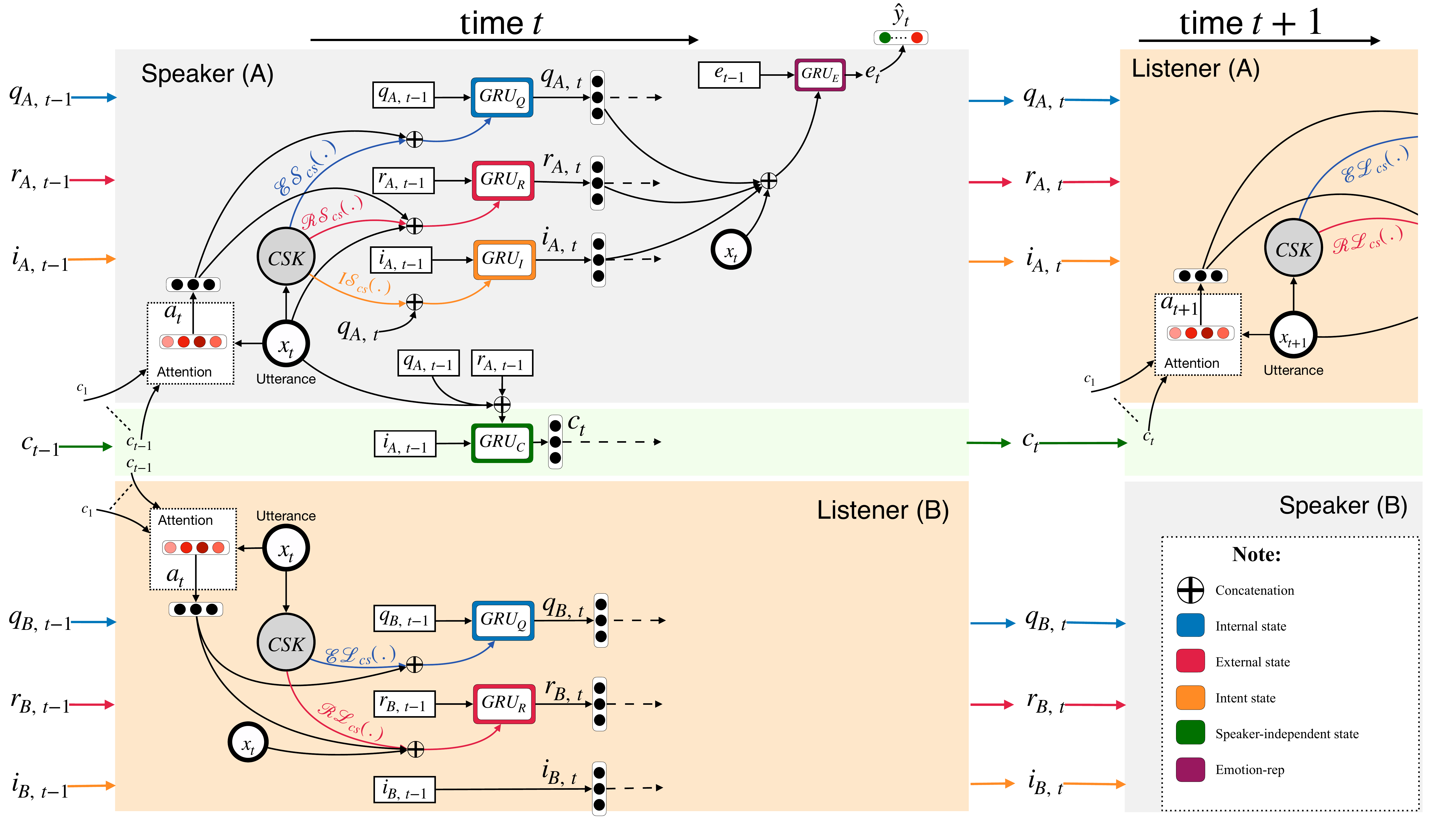}
    \caption{Illustration of {\sc Cosmic} framework. \textit{CSK}: Commonsense knowledge from COMET. In practice we use Bidirectional GRU cells. However, for clarity unidirectional cells are shown in the sketch.}
    \label{fig:framework}
\end{figure*}

The overall architecture of the {\sc Cosmic} framework is illustrated in Figure \ref{fig:framework}.




\subsection{Context Independent Feature Extraction}
\label{sec:text-feat-extr}

We employ the RoBERTa model \cite{liu2019roberta} to extract context independent utterance level feature vectors.
We first fine-tune the RoBERTa Large model for emotion label prediction from the transcript of the utterances. RoBERTa Large follows the original BERT Large \cite{devlin2018bert} architecture having 24 layers, 16 self-attention heads in each block and a hidden dimension of 1024, resulting in a total of 355M parameters. Let an utterance $x$ consists of a sequence of BPE tokenized tokens $x_1, x_2, \dots, x_N$, with emotion label $E_x$. In this setting, the fine-tuning of the pretrained RoBERTa model is realized through a sentence classification task. A special token $[CLS]$ is appended at the beginning of the utterance to create the input sequence for the model: $[CLS], x_1, x_2, \dots, x_N$. This sequence is passed through the model, and the activation from the last layer corresponding to the $[CLS]$ token is then used in a small feedforward network to classify it into its emotion class $E_x$.

Once the model has been fine-tuned for emotion label classification, we pass the $[CLS]$ appended BPE tokenized utterances to it and extract out activations from the final four layers corresponding to the $[CLS]$ token. These four vectors are then averaged to obtain the context independent utterance feature vector with a dimension of 1024.

\subsection{Commonsense Feature Extraction}
\label{sec:csk-feat-extr}

\begin{table}[ht]
\small
 	\centering
 	\resizebox{\linewidth}{!}{
	\begin{tabular}{C{2.6cm}ccC{1.25cm}}
		\toprule
		Commonsense Feature & Notation & Nature & Causal Relation \\
		\midrule
		Intent of speaker & $\mathcal{IS}_{cs}(.)$ &  Mental state & Cause\\
		Effect on speaker & $\mathcal{ES}_{cs}(.)$ &  Mental state & Effect \\
		Reaction of speaker & $\mathcal{RS}_{cs}(.)$ & Event & Effect \\
		Effect of listeners & $\mathcal{EL}_{cs}(.)$ & Mental state & Effect \\
		Reaction of listeners & $\mathcal{RL}_{cs}(.)$ & Event & Effect \\
		\bottomrule
	\end{tabular}
	}
	\caption{Functional notations of commonsense knowledge used in {\sc Comet}. The functions take as input the utterance $u$ and returns the feature indicated in the leftmost column. Intent and effect on speaker and listeners can be categorized into \textit{mental states}, whereas their reactions are \textit{events}. Intent is also a \textit{causal} variable whereas the rest are \textit{effects}.\label{tab:csk}}
\end{table}

In this work, we use the commonsense transformer model COMET \cite{bosselut2019comet} to extract the commonsense features. COMET is trained on several commonsense knowledge graphs to perform automatic knowledge base construction. The model is given a triplet \{$s, r, o$\} from the graph and is trained to generate the object phrase $o$ from concatenated subject phrase $s$ and relation phrase $r$. COMET is an encoder-decoder model that uses the pretrained autoregressive language model GPT \cite{radford2018improving} as the base generative model.

To perform the task of generative commonsense knowledge construction, COMET is trained on ATOMIC (The Atlas of Machine Commonsense) \cite{sap2019atomic}, a collection of everyday inferential \textit{if-then} commonsense knowledge organized through textual descriptions.
ATOMIC consists of nine different if-then relation types to distinguish agents vs themes, causes vs effects, voluntary vs non-voluntary events, and actions vs mental states. Given an event in which X participates, the nine relation types ($r$) are inferred as follows: i) \textit{intent of X}, ii) \textit{need of X}, iii) \textit{attribute of X}, iv) \textit{effect on X}, v) \textit{wanted by X}, vi) \textit{reaction of X}, vii) \textit{effect on others}, viii) \textit{wanted by others}, and ix) \textit{reaction of others}.
As an example, given an event or subject phrase ($s$): “Person X gives Person Y a compliment”, the inference from COMET for relation phrase ($r$): \textit{intent of X} and \textit{reaction of others} would be “X wanted to be nice” and “Y will feel flattered” respectively.

COMET is a generative model and as illustrated in the above example it produces a discrete sequence of commonsense knowledge conditioned on the subject and relation phrase. In our model however, we make use of continuous vectors of commonsense representations. For that, we take the pretrained COMET model on ATOMIC knowledge graph and discard the phrase generating decoder module. We treat utterance $U$ as the subject phrase and concatenate it with the relation phrase $r$. Next, we pass the concatenated $\{U \oplus r\}$ through the encoder of COMET and extract out the activations from the final time-step. In particular we use the relations presented in Table \ref{tab:csk}: \textit{intent of X}, \textit{effect on X}, \textit{reaction of X}, \textit{effect on others} and \textit{reaction of others} (where $X$ is the speaker and $others$ are listeners). Performing this feature extraction operation results in five different vectors (respective to the five different relations) for each utterance in the conversation. These vectors are 768 dimensional.

The nature of the various relation types in ATOMIC allows us to extend it naturally to conversational frameworks. The relations enable the modeling of phenomenons such as content (event, persona, mental states) and causal relations (cause, effect, stative) which are essential elements for understanding conversational context. These different relations are of key importance because generally there is a major interplay between virtually all of them throughout the course of a conversation. For instance, the relations i) - vi) are all intrinsically related to the speaker and vii) - ix) are all akin to the listener. On a more fine-grained level, the \textit{intent, effect} and \textit{react} components of the speaker and listener are all elemental for understanding the nature of the conversation. We surmise that adopting these relational variables in a unified framework would be highly useful to create enhanced representations of the conversation.

\subsection{Commonsense Conversational Model}
\label{sec:model}

We first introduce our notations and present a high level view of the main architecture of our {\sc Cosmic} model. A conversation consists of $N$ utterances $u_1, u_2, \dots, u_N$,  in which $M$ distinct speakers/participants $p_1, p_2, \dots, p_M$ take part. Utterance $u_t$ is spoken by participant $p_{s(u_t)}$. For every $t \in \{1,2, \dots, N\}$, we denote context independent RoBERTa vectors by $x_t$. Commonsense vectors corresponding to \textit{intent of X}, \textit{effect on X, reaction of X, effect on others} and \textit{reaction of others} are denoted by
$\mathcal{IS}_{cs}(u_t), \mathcal{ES}_{cs}(u_t), \mathcal{RS}_{cs}(u_t), \mathcal{EL}_{cs}(u_t),$ and $\mathcal{RL}_{cs}(u_t)$
respectively. $X$ is assumed to be the speaker and $others$ are assumed to be the listeners.

Since conversations are highly sequential in nature and contextual information flows along a sequence, a \textit{context state} $c_t$ and \textit{attention vector} $a_t$ are formulated that model the sequential dependency between utterances. The \textit{context state} and \textit{attention vector} are always shared between all the participants of the conversation.

\begin{table}[t]
\small
\begin{center}
 	\centering
	\begin{tabular}{C{2.5cm}C{4cm}}
		\toprule
		State & Influenced By \\
		\midrule
		\multirow{2}{*}{Context State} & Utterance, \\
		& Internal state, External state \\
		\hline
		\multirow{2}{*}{Internal State} & Context state,  \\
		& \textit{Effect on speaker, listener} \\
		\hline
		\multirow{2}{*}{External State} & Context state, Utterance, \\
		& \textit{Reaction of speaker, listener} \\
		\hline
		\multirow{1}{*}{Intent State} & Internal state, \textit{Intent of speaker}\\
		\hline
		\multirow{2}{*}{Emotion State} & Utterance, Intent state\\
		& Internal state, External state \\
		\bottomrule
	\end{tabular}
	\caption{Different states and the respective variables they are influenced by. \textit{Italic} variables are forms of commonsense knowledge from Table \ref{tab:csk}. \label{tab:states}}
\end{center}
\end{table}

An \textit{internal state}, \textit{external state} and \textit{intent state} are used to model different mental states, actions and events for the participants. These are represented by $q_{k,t}, r_{k,t}$ and $i_{k,t}$ for the participants $k \in [1,2,\dots, M]$. The \textit{internal state} and the \textit{external state} can be collectively considered as the speaker state. This states are necessary to capture the complex mental and emotional dynamics of the participants. The emotion state $e_t$ is then modelled from a combination of the three states and the immediate preceding emotion state. Finally the appropriate emotion class for the utterance is inferred from the emotion state.

In our framework, context and commonsense modeling is performed using GRU cells \cite{chung2014empirical}. GRU cells take as input $y_t$ and update its hidden state from $h_{t-1}$ to $h_t$ using the transformation: $h_t = GRU(h_{t-1}, y_t)$. New hidden state $h_t$ also serves as the output of the current step. The cell is parameterized by weights $W$ and biases $b$ of appropriate sizes depending upon the
input $y_t$ and output $h_t$.
We use five Bidirectional GRU cells $GRU_C, GRU_Q, GRU_R, GRU_I$, and $GRU_E$ for modeling context state, internal state, external state, intent state, and emotion state respectively. For ease of representation we formulate the different states with unidirectional GRU cells here.

\paragraph{Context State:}
The context state stores and propagates the overall utterance-level information
along the sequence of the conversation flow. This state is updated using context GRU cell $GRU_C$ after each time-step $t$ when the utterance is uttered by some participant $p_{s(u_t)}$. RoBERTa feature vector $x_t$, internal state $q_{s(u_t), t-1}$, and external state $r_{s(u_t), t-1}$ of the speaker from the immediate previous time-step (just before uttering the utterance) are concatenated and serve as the input vector for $GRU_C$.
\begin{equation}
c_{t} = GRU_C(c_{t-1}, (x_t \oplus q_{s(u_t), t-1} \oplus r_{s(u_t), t-1}))
\label{eq:0}
\end{equation}
We also pool attention vector $a_t$ from the history of context $[c_1, c_2, \dots, c_{t-1}]$ using soft-attention. This attention vector is later used to perform updates on internal and external states.
\begin{align}
& u_i = tanh(W_s c_i + b_s), \hspace{2px}i \in [1, t-1]  \nonumber\\
& \alpha_i = \frac{exp(u_i^{T}x_i)}{\sum_{i=1}^{t-1}{exp(u_i^{T}x_i)}} \nonumber\\
& a_t = \sum_{i=1}^{t-1}{\alpha_i c_i}
\label{eq:1}
\end{align}

\paragraph{Internal State:}
The internal state of the participants is conditioned on how the individual is feeling and what is the effect perceived from other participants.
This state may remain concealed,  as participants may not always express explicitly their feeling or outlook through external stance or reactions. Apart from feelings, this state can also be considered to include aspects that the participant actively tries not to express or features that are considered common knowledge and don't require explicit communication. The effect on oneself is thus elemental to represent the internal state of the participants. We model the internal state of the participants using $GRU_Q$. For time-step $t$, the internal state of the speaker $p_{s(u_t)}$ is updated by taking into account the attention vector $a_t$ and commonsense vector \textit{effect on speaker} $\mathcal{ES}_{cs}(u_t)$
\begin{equation}
q_{s(u_t), t} = GRU_Q(q_{s(u_t), t-1}, (a_t \oplus \mathcal{ES}_{cs}(u_t))) \label{eq:2}
\end{equation}
For all the other participants apart from the speaker, this update is performed using \textit{effect on listeners} $\mathcal{EL}_{cs}(u_t)$.
\begin{equation}
q_{j, t} = GRU_Q(q_{j, t-1}, (a_t \oplus \mathcal{EL}_{cs}(u_t))); \forall j \neq s(u_t) \label{eq:3}
\end{equation}

\paragraph{External State:}
Unlike the internal state, the external state of the participants is all about the expressions, reactions, and responses. Naturally, this state can be easily seen, felt, or understood by the other participants. For instance, the actual utterance, the manner of articulation, the speech, and other acoustic features, the visual expression, gestures, and stance can all be loosely considered to fall under the regime of external state. $GRU_R$ updates the external state of the speaker $p_{s(u_t)}$ by taking as input the concatenation of attention vector $a_t$, utterance vector $x_t$ and commonsense vector \textit{reaction of speaker} $\mathcal{RS}_{cs}(u_t)$
\begin{equation}
r_{s(u_t), t} = GRU_R(r_{s(u_t), t-1}, (a_t \oplus x_t \oplus \mathcal{RS}_{cs}(u_t))) \label{eq:4}
\end{equation}
For listeners, this update is performed using \textit{reaction of listeners} $\mathcal{RL}_{cs}(u_t)$.
\begin{equation}
\begin{aligned}
r_{j, t} &= GRU_R(r_{j, t-1}, (a_t \oplus x_t \oplus & \mathcal{RL}_{cs}(u_t))); \\
\forall j &\neq s(u_t)
\label{eq:5}
\end{aligned}
\end{equation}

\paragraph{Intent State:}
Intent is a mental state that represents the commitment to carry out a particular set of actions. The intent of the speaker always plays a crucial role in determining the emotional dynamics of a conversation. The intent of the speaker changes from $i_{s(u_t), t-1}$ to $i_{s(u_t), t}$ at time-step $t$. This change is invoked by the commonsense \textit{intent of speaker} vector $\mathcal{IS}_{cs}(u_t)$ and internal speaker state $q_{s(u_t), t}$ at that respective time-step $t$. The intent states are captured by GRU cell $GRU_I$:
\begin{equation}
i_{s(u_t), t} = GRU_I(i_{s(u_t), t-1}, (\mathcal{IS}_{cs}(u_t) \oplus q_{s(u_t), t})) \label{eq:6}\\
\end{equation}
The intent of the listener(s), however, is kept unchanged. This is because the intent of a participant who is silent should not change. The change should occur only when the particular participant speaks again.
\begin{equation}
i_{j, t} = i_{j, t-1}; \forall j \neq s(u_t) \label{eq:7}
\end{equation}

\paragraph{Emotion State:}
The emotional state determines the emotional mood of the speaker and the emotion class of the utterance. We posit that the emotional state depends upon the utterance and composite state of the speaker that takes into account the internal, external, and intent state. Naturally the current emotion state also depends on the previous emotion state of the speaker. $GRU_E$ captures the emotion state by combining all of the factors as following,
\begin{equation}
e_{t} = GRU_E(e_{t-1}, (x_t \oplus q_{s(u_t), t} \oplus r_{s(u_t), t} \oplus i_{s(u_t), t})) \label{eq:8}
\end{equation}

\paragraph{Emotion Classification:} Finally all the utterances in the conversation are classified with a fully connected network from $e_t$
\begin{align}
    & P_t = softmax(W_{smax}e_t + b_{smax}); \forall t \in [1, N]\nonumber \\
    & \hat{y_t}=\argmax_{k}(\mathcal{P}_t[k])
\end{align}

\section{Experimental Setup}
\label{sec:exp}
\subsection{Datasets}
\begin{table}[ht!]
\centering
\resizebox{\linewidth}{!}{
        \begin{tabular}{L{20mm}|c|c|c|c|c|c}
            \toprule
            \multirow{2}{*}{Dataset}&\multicolumn{3}{c|}{$\#$ dialogues}&\multicolumn{3}{c}{$\#$ utterances}\\
            &train&val&test&train&val&test\\
            \hline \hline
            IEMOCAP &120&12 & 31 & \multicolumn{2}{c|}{5810} & 1623\\
            DailyDialog & 11,118 & 1,000 & 1,000 & 87,832 & 7,912 & 7,863\\
            MELD & 1,039 & 114 & 280 & 9,989 & 1,109 & 2,610\\
            EmoryNLP & 659 & 89 & 79 & 7,551 & 954 & 984\\
            \bottomrule
            \multicolumn{7}{c}{}
        \end{tabular}
        }
\resizebox{\linewidth}{!}{
        \begin{tabular}{L{20.4mm}|c|c}
            \toprule
            Dataset & {$\#$ classes} & Metric \\
            \hline \hline
            IEMOCAP & 6 & Weighted Avg. F1 \\
            DailyDialog & 7* & Macro F1 and Micro F1\\
            MELD & 3 and 7 & Weighted Avg. F1 over 3 and 7 classes \\
            EmoryNLP & 3 and 7 & Weighted Avg. F1 over 3 and 7 classes\\
            \bottomrule
        \end{tabular}
        }
    \caption{Statistics of splits
    and evaluation metrics used in different datasets.
    In MELD and EmoryNLP evaluation is performed for 3 class (broad) and 7 class (fine-grained) classification. \textit{Neutral}* classes constitutes to 83\% of the DailyDialog dataset. These are excluded when calculating the Micro F1 score.}
    \label{table:data}
\end{table}

We benchmark {\sc Cosmic} on four different conversational emotion recognition datasets: i) \textbf{IEMOCAP} \cite{iemocap} ii) \textbf{MELD} \cite{poria2018meld} iii) \textbf{DailyDialog} \cite{li2017dailydialog}, and iv) \textbf{EmoryNLP} \cite{zahiri2018emotion}. IEMOCAP and DailyDialog are two-party datasets, whereas MELD and EmoryNLP are multi-party datasets.
We report experimental results for conversational emotion recognition from the textual information for all four datasets.
Information about the datasets is shown in Table \ref{table:data}.

\textbf{IEMOCAP}~\cite{iemocap} is a dataset of two person conversations among ten different unique speakers. The train set dialogues come from the first eight speakers, whereas the test set dialogues are from the last two. Each utterance is annotated with one of the following six emotions: \textit{happy, sad, neutral, angry, excited,} and \textit{frustrated}.

\textbf{DailyDialog}~\cite{li2017dailydialog}
covers various topics about our daily life and follows the natural human communication approach. All utterances are labeled with both emotion categories and dialogue acts. The emotion can belong to one of the following seven labels: \textit{anger, disgust, fear, joy, neutral, sadness}, and \textit{surprise}. The dataset has over 83\% \textit{neutral} labels and these are excluded during Micro-F1 evaluation.

\textbf{MELD}~\cite{poria2018meld} is
a multimodal dataset extended from the EmotionLines dataset \cite{chen2018emotionlines}. MELD is
collected from the TV show \textit{Friends} and has more than 1400 dialogues and 13000 utterances. Utterances
are labeled with emotion and sentiment classes. The emotion classes  belong to \textit{anger, disgust, sadness, joy, surprise, fear}, or \textit{neutral}, and the sentiment classes  belong to \textit{positive, negative} or \textit{neutral}.

\textbf{EmoryNLP}~\cite{zahiri2018emotion} is another dataset also based on the show \textit{Friends}. Utterances in this dataset are annotated on seven and three emotion classes. The seven emotion classes are \textit{neutral, joyful, peaceful, powerful, scared, mad} and \textit{sad}. To create three emotion classes: \textit{joyful, peaceful}, and \textit{powerful} are grouped together to form the \textit{positive} class; \textit{scared, mad} and \textit{sad} are grouped together to form the \textit{negative} class; and the \textit{neutral} class is kept unchanged.

\subsection{Training Setup}
For context independent feature extraction, the RoBERTa model is fine-tuned on the set of all utterances and their emotion labels in the training data. We fine-tune the RoBERTa model for a batch size of 32 utterances with Adam optimizer with learning rate of 1e-5. In the case of MELD and EmoryNLP datasets, we use a residual connection between the first and the penultimate layer which brings more stability in the training in the emotion recognition model.
The emotion recognition model is trained with Adam optimizer having a learning rate of 1e-4.

\begin{table*}[ht!]
  \centering
  \resizebox{\linewidth}{!}{
   \begin{tabular}{l|L{4cm}||C{1.4cm}||C{1.2cm}C{1.25cm}||C{1.5cm}C{1.5cm}||C{1.5cm}C{1.5cm}}
    \toprule
   & \multirow{3}{*}{Methods} & \multicolumn{1}{c||}{IEMOCAP} & \multicolumn{2}{c||}{DailyDialog} & \multicolumn{2}{c||}{MELD} & \multicolumn{2}{c}{EmoryNLP}\\
    \cline{3-9} & & \multirow{2}{*}{W-Avg F1} & \multirow{2}{*}{Macro F1} & \multirow{2}{*}{Micro F1} & W-Avg F1 (3-cls) & W-Avg F1 (7-cls) & W-Avg F1 (3-cls) & W-Avg F1 (7-cls) \\
    \toprule
   \multirow{5}{*}{\rot{GloVe-based}}& CNN & 52.04 & 36.87 & 50.32 & 64.25 & 55.02 & 38.05 & 32.59 \\
  &  ICON & 58.54 & - & - & - & - & - & -\\
  &  KET & 59.56 & - & 53.37 & - & 58.18 & - & 34.39 \\
  &  ConGCN & - & - & - & - & 57.40 & - & - \\
  &  DialogueRNN & 62.57 & 41.80 & 55.95 & 66.10 & 57.03 & 48.93 & 31.70 \\
    \midrule
\multirow{8}{*}{\rot{(Ro)BERT(a)-based}}  &  BERT DCR-Net & - & 48.90 & - & - & - & - & -\\
  &  BERT+MTL & - & - & - & - & 61.90 & - & 35.92 \\
 &  RoBERTa & 54.55 & 48.20 & 55.16 & 72.12 & 62.02 & 55.28 & 37.29 \\
  &  RoBERTa DialogueRNN & 64.76 & 49.65 & 57.32 & 72.14 & 63.61 & 55.36 & 37.44\\

    \cline{2-9}
  &  \textbf{COSMIC} & \textbf{65.28} & \textbf{51.05} & \textbf{58.48} & \textbf{73.20} & \textbf{65.21} & \textbf{56.51} & \textbf{38.11}\\
 &   \quad  \footnotesize{w/o Speaker CSK} & 63.27 & 50.18 & 57.45 & 72.94 & 64.41 & 55.46 & 37.35 \\
 &   \quad  \footnotesize{w/o Listener CSK}& 65.05 & 48.67 & 58.28 & 72.90 & 64.76 & \textbf{56.57} & \textbf{38.15} \\
  &  \quad  \footnotesize{w/o Speaker, Listener CSK} & 63.05 & 48.68 & 56.16 & 72.62 & 64.28 & 55.34 & 37.10 \\
    \bottomrule
   \end{tabular}
  }
  \caption{Comparison of results against various
  methods. Scores are average of five runs. Test scores are computed at best validation scores. {\sc Cosmic} achieves new state-of-the-art results across all the datasets. 
  CSK refers to commonsense knowledge components from COMET. We report the average score of the 10 runs for RoBERTa DialogueRNN and COSMIC. The CNN and DialogueRNN scores using Glove embeddings are obtained from ~\cite{ghosal2020utterance}.}
  \label{tab:results1}
\end{table*}

\section{Results and Analysis}
\label{sec:result}

\subsection{Baseline and State-of-the art Methods}
For a comprehensive evaluation of {\sc Cosmic}, we compare it against the following methods: \textbf{CNN}~\cite{kim2014convolutional} is a convolutional neural network model trained on top of pretrained GloVe embeddings. Standard configurations of filter sizes are used.
The model is trained at the utterance level to predict the emotion classes.
\textbf{ICON}~\cite{hazarika-EtAl:2018:N18-1} uses two GRU networks to learn the utterance representations for dialogues between two-participants. The output of the two speaker GRUs is then connected using another GRU that helps in performing explicit inter-speaker modeling.
ICON is limited to conversations with only two participants only.
\textbf{KET}~\cite{zhong2019knowledge} or Knowledge enriched transformers dynamically leverages external commonsense knowledge using hierarchical self-attention and context aware graph attention.
\textbf{ConGCN}~\cite{zhang2019modeling} considers utterances and participants of a conversation as nodes of graph network and models both context and speaker sensitive dependence for emotion detection.
\textbf{BERT DCR-Net}~\cite{qindcr2020dcrnet} is a deep co-interactive relation network that uses BERT based features for joint dialogue act recognition and emotion (sentiment) classification.
A relation layer learns to explicitly model the relation and interaction between these two tasks in a multi-task setting.
\textbf{BERT+MTL}~\cite{Li2020MultiTaskLW} is a multi-task learning framework where features extracted from BERT are used in a recurrent neural network for emotion recognition and speaker identification.
\textbf{DialogueRNN}~\cite{dialoguernn} models the emotion of utterances
in a conversation with
speaker, context and emotion information from neighbour utterances.
These factors are modeled using three separate GRU networks to  keep track of the individual speaker states.

We report and compare the performance of {\sc Cosmic} on test data
in Table \ref{tab:results1}.
State-of-the-art models 
use GloVe embeddings to extract context-independent features. As features extracted from transformer based networks such as BERT and RoBERTa generally outperform traditional word embeddings such as word2vec and GloVe, we also report results of the models when used with
BERT or RoBERTa features.
\begin{figure*}[t]
    \centering
    \includegraphics[width=\linewidth]{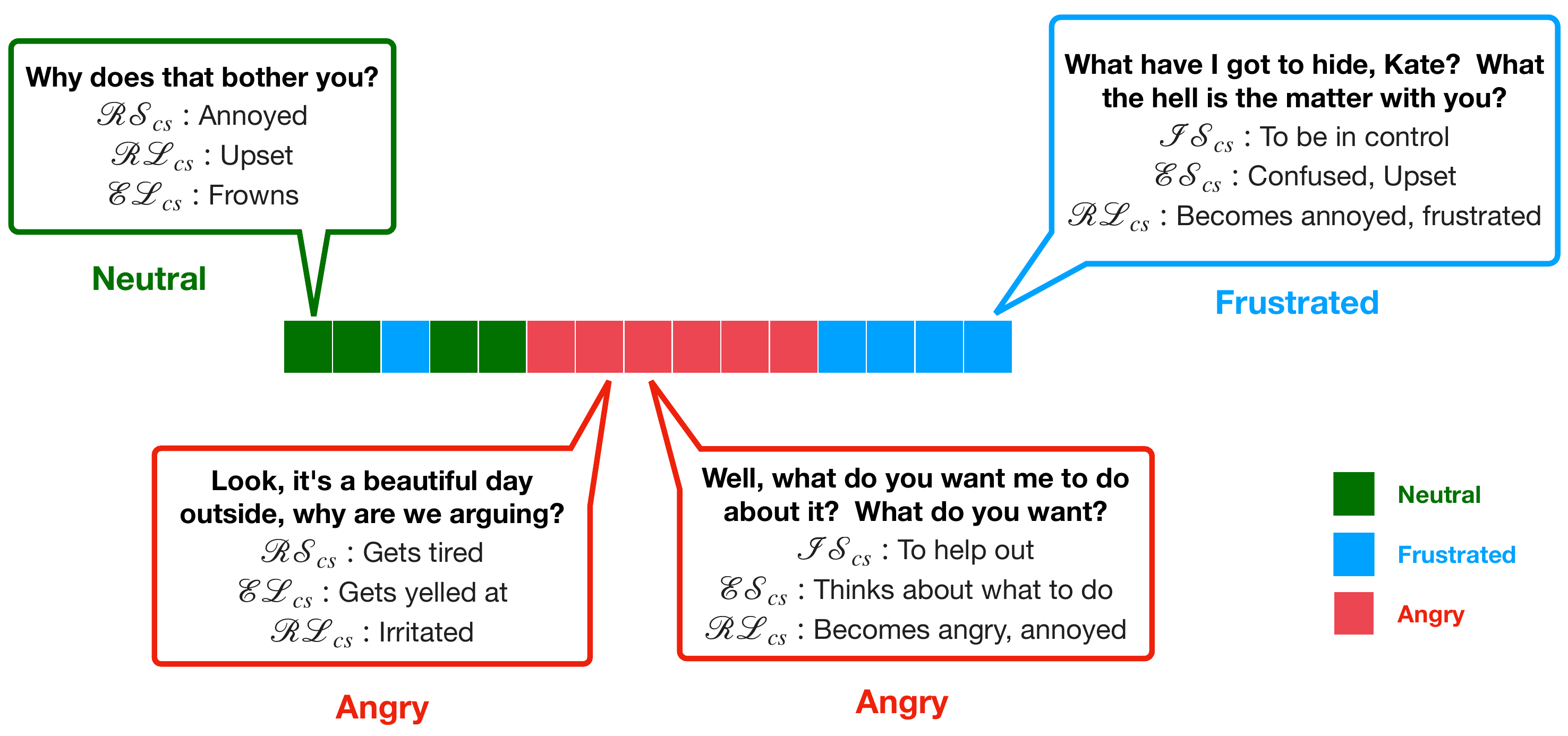}
    \caption{Case study from the IEMOCAP dataset. Discrete commonsense sequences are shown for more interpretability. Commonsense knowledge helps in predicting emotion shifts and understanding difference between closely related emotion classes such as \textit{angry} and \textit{frustrated}.}
    \label{fig:case_study}
\end{figure*}
\subsection{Comparison with the State-of-the-Art Methods}
\paragraph{IEMOCAP and DailyDialog:} IEMOCAP and DailyDialog contain dyadic conversations with mostly natural and coherent utterances. We observe that RoBERTa features improve the DialogueRNN 
models, and other BERT based models perform similarly.
{\sc Cosmic} improves over all the models, however the improvement on IEMOCAP is not as large as it is on DailyDialog. {\sc Cosmic} achieves new state-of-the-art scores of 65.28 on IEMOCAP; 51.05 and 58.48 in DailyDialog for the two different evaluation metrics.

\paragraph{MELD and EmoryNLP:} These two datasets have been annotated from the TV show Friends, and  utterances are often very short. Although dialogues occasionally contain emotion specific words, this does not happen very often at the utterance level. Naturally, emotion dynamics are highly contextual in nature and almost always depend on surrounding utterances. It has been observed in previous work that emotion modeling in MELD is difficult because often there are a lot of speakers in each conversation but they utter only a small number of utterances. 
 Sophisticated models such as DialogueRNN 
 do not bring as much improvement over CNN as they do on IEMOCAP.
We observe that, {\sc Cosmic} brings a large improvement over other models on the fine-grained (7 class) classification setup for both  datasets. It achieves new state-of-the-art weighted F1 scores of 73.20 and 56.51 on three class classification; 65.21 and 38.11 on seven class classification on MELD and EmoryNLP.

\subsection{The Role of Commonsense}
In Table \ref{tab:results1}, we also report results of ablation studies by removing listener-specific and speaker-specific commonsense components. For speaker ablation, we discard $\mathcal{IS}_{cs}(u_t), \mathcal{ES}_{cs}(u_t), \mathcal{RS}_{cs}(u_t)$, and observe a sharp drop in performance in most cases. For listener ablation, we discard $\mathcal{EL}_{cs}(u_t),$ and $\mathcal{RL}_{cs}(u_t)$ and find that the performance also drops but not as much as the speaker ablation. In fact, listener ablation leads to slight improvement in performance in EmoryNLP. The results suggest that speaker-specific commonsense has a greater impact in the overall performance of {\sc Cosmic}, which is expected because we are predicting the emotion class of the speaker at each utterance. Finally, ablation with respect to both components at the same time naturally leads to higher drop in overall performance.

\subsection{Case Study}
We illustrate a case study on a test conversation instance from the IEMOCAP dataset in Figure \ref{fig:case_study}. The conversation begins with a couple of \textit{neutral} utterances, but then the situation quickly escalates, and finally, it ends with a lot of \textit{angry} and \textit{frustrated} utterances from both the speakers. State-of-the-art models like DialogueRNN 
often find this kind of scenarios difficult, when there is a couple of sudden emotions shifts in between (\textit{neutral} to \textit{frustrated} and then \textit{neutral} again). These models also tend to misclassify utterances that have subtle differences in emotion classes such as \textit{frustrated} and \textit{angry}. In {\sc Cosmic}, the propagation of commonsense knowledge makes it easier for the model to handle the sudden transitions and to understand the subtle difference between closely related emotion classes. In Figure \ref{fig:case_study}, for the first utterance, the commonsense model predicts that the \textit{reaction of speaker} is \textit{annoyed} and propagation of this information helps in predicting that the speaker's next utterance actually belongs to the \textit{frustrated} class. Similarly for the rest of the illustrated utterances, the commonsense knowledge from \textit{effect on speaker} and \textit{reaction of listener} helps the model in distinguishing and predicting the \textit{anger} and \textit{frustrated} classes correctly.

\subsection{Strategies to Incorporate Commonsense} Apart from the five commonsense features that we use in {\sc Cosmic} (Table \ref{tab:csk}), there are four other features that can be extracted from COMET: \textit{attribute of speaker}, \textit{need of speaker}, \textit{wanted by speaker}, and \textit{wanted by listeners}. We incorporate them using different strategies that  add extra complexity in our framework but ultimately do not improve the performance by a significant margin. We experimented along the following directions:

$\bullet$ \textit{Attribute of speaker} is loosely considered as a personality trait.
This latent variable influenced the \textit{internal, external} and \textit{intent states}. We find that the discrete \textit{attribute} features from COMET are mostly a single word like `stubborn', `patient', `argumentative', `calm', etc and they change quite abruptly for the same participant in continuing utterances. Hence, we find that their vectorized representations do not help much.

$\bullet$ \textit{Need of speaker}, \textit{wanted by speaker}, and \textit{wanted by listeners} are considered as output variables that are to be predicted from the input utterance and the five basic commonsense features (Table \ref{tab:csk}). We add auxiliary output functions and jointly optimize the emotion classification loss with mean-squared loss between predictions and reference commonsense vectors. This strategy also does not help much in improving the emotion classification performance.

Although the performance improvement is observed using commonsense knowledge across the datasets, this improvement is not very substantial. In the future, we plan to identify better commonsense knowledge sources and develop models that can infuse this knowledge into deep learning models more efficiently.



\section{Conclusion}
In this work, we presented {\sc Cosmic}, a framework that models various aspects of commonsense knowledge by considering mental states, events, actions, and cause-effect relations for emotion recognition in conversations. Using commonsense representations, our model alleviates issues such as difficulty in detecting emotion shifts and misclassification between related emotion classes that are often present in current RNN and GCN based methods. {\sc Cosmic} achieves new state-of-the-art results for emotion recognition across several benchmark datasets.

 \section*{Acknowledgements}
 This research is supported by A*STAR under its RIE 2020 AME 
 programmatic grant RGAST2003, 
 by the National Science Foundation (grant \#1815291), and by the John Templeton Foundation (grant \#61156). Any opinions, findings, and conclusions or recommendations expressed in this material are those of the authors and do not necessarily reflect the views of A*STAR, the National Science Foundation, or the John Templeton Foundation.

\bibliography{emnlp2020}
\bibliographystyle{acl_natbib}

\end{document}